\documentclass{article} % For LaTeX2e
\usepackage{iclr2017_conference,times}
\usepackage{url}
\usepackage{graphicx}
\usepackage{subcaption}
\title{Compact Deep Convolutional Neural \\Networks with Coarse Pruning}

\author{Sajid Anwar,  Wonyong Sung\\
Department of Electrical Engineering and Computer Science\\
Seoul National University\\
Gwanak-Gu, 08826, Republic of Korea \\
\texttt{sajid@dsp.snu.ac.kr, wysung@snu.ac.kr} \\
}

\begin{document}

\maketitle

\begin{abstract}
The learning capability of a neural network improves with increasing depth at higher computational costs. Wider layers with dense kernel connectivity patterns furhter increase this cost and may hinder real-time inference. We propose feature map and kernel level pruning for reducing the computational complexity of a deep convolutional neural network. Pruning feature maps reduces the width of a layer and hence does not need any sparse representation. Further, kernel pruning converts the dense connectivity pattern into a sparse one. Due to coarse nature, these pruning granularities can be exploited by GPUs and VLSI based implementations. We propose a simple and generic strategy to choose the least adversarial pruning masks for both granularities. The pruned networks are retrained which compensates the loss in accuracy. We obtain the best pruning ratios when we prune a network with both granularities. Experiments with the CIFAR-10 dataset show that more than 85\% sparsity can be induced in the convolution layers with less than 1\% increase in the missclassification rate of the baseline network. 
\end{abstract}

\section{Introduction}
\label{sec_introduction}

Deep and wider neural networks have the capacity to learn a complex unknown function from the training data. The network reported in \cite{dean2012large} has 1.7 Billion parameters and is trained on tens of thousands of CPU cores. Similarly \cite{simonyan2014very} has employed 16-18 layers and achieved excellent classification results on the ImageNet dataset. The high computationally complexity of wide and deep networks is a major obstacle in porting the benefits of deep learning to resource limited devices. Therefore, many researchers have proposed ideas to accelerate deep networks for real-time inference \cite{yu2012exploiting, han2015learning, han2015deep, mathieu2013fast, anwar2015structured}.  

Network pruning is one promising techique that first learns a function with a suficiently large sized network followed by removing less important connections \cite{yu2012exploiting, han2015learning, anwar2015structured}. This enables smaller networks to inherit knowledge from the large sized predecessor networks and exhibit comparable level of performance. The works of \cite{han2015learning,han2015deep} introduce fine grained sparsity in a network by pruning scalar weights. Due to unstructured sparsity, the authors employ compressed sparse row/column (CSR/CSC) for sparse representation. Thus the fine grained irregular sparsity cannot be easily translated into computational speedups.

\begin{figure*}
\centering
\includegraphics[width=\textwidth]{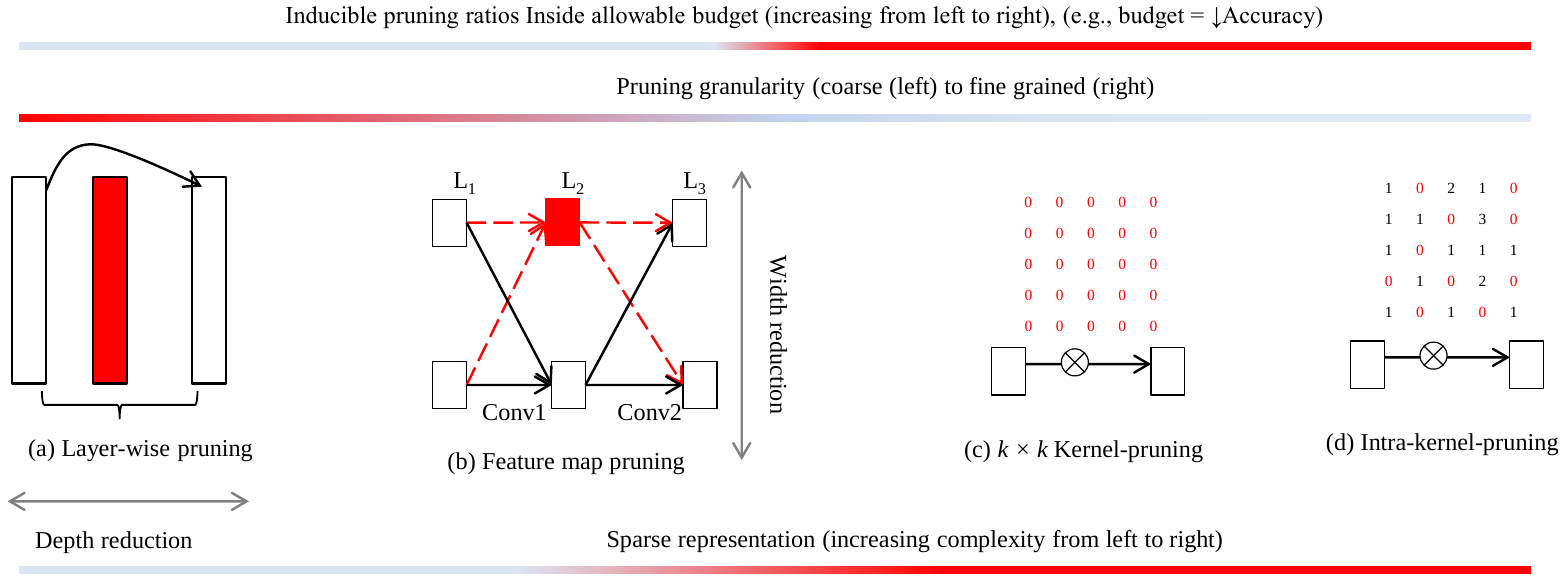}
\caption{(a-d) shows four possible pruning granularities. The proposed work is focussed on the (b) feature map and (c) kernel pruning for simple sparse represenation. It can be observed that for the depicted architecture in Fig. (b), four convolution kernels are pruned.}
\label{layer_fmap_kernel_intrakernel}
\end{figure*} 

Sparsity in a deep convolutional neural network (CNN) can be induced at various levels. Figure~\ref{layer_fmap_kernel_intrakernel} shows four pruning granularities. At the coarsest level, a full hidden layer can be pruned. This is shown with a red colored rectangle in Fig. \ref{layer_fmap_kernel_intrakernel}(a). Layer wise pruning affects the depth of the network and a deep network can be converted into a shallow network. Increasing the depth improves the network performance and layer-wise pruning therefore demand intelligent techniques to mitigate the performance degradation. The next pruning granularity is removing feature maps \cite{polyak2015channel, anwar2015structured}. Feature map pruning removes a large number of kernels and is therefore very destructive in nature. We therefore may not achieve higher pruning ratios. For the depicted architecture in Fig. \ref{layer_fmap_kernel_intrakernel} (b)., pruning a single feature map, zeroes four kernels. Feature map pruning affects the layer width and we directly obtain a thinner network and no sparse representation is needed. Kernel purning is the next pruning granularity and it prunes $k \times k$ kernels. It is neither too fine nor too coarse and is shown in Fig. \ref{layer_fmap_kernel_intrakernel}(c). Kernel pruning is therefore a balanced choice and it can change the dense kernel connectivity pattern to sparse one. Each convolution connection involves $W~\times~H~\times~k~\times~k$ MAC operations where \textit{W, H} and \textit{k} represents the feature map width, height and the kernel size respectively. Further the sparse representation for kernel-Pruning is also very simple. A single flag is enough to represent one convolution connection.The conventional pruning techniques induce sparsity at the finest granularity by zeroing scalar weights. This sparsity can be induced in much higher rates but demands sparse representation for computational benefits \cite{han2015learning}. Therefore the high pruning ratios do not directly translate into computational speedups. Figure~\ref{layer_fmap_kernel_intrakernel}(d) shows this with red colored zeroes in the kernel.  Further Fig.~\ref{layer_fmap_kernel_intrakernel} summarizes the relationship between three related factors: the pruning granularities, the pruning ratios and the sparse representations. Coarse pruning granularities demand very simple sparse representation but we cannot achieve very high pruning ratios. Similarly fine grained pruning granularities can achieve higher pruning ratios but the sparse representation is more complicated. 

The reference work of \cite{anwar2015structured} analysed feature map pruning with intra-kernel strided sparsity. To reduce the size of feature map and kernel matrices, they further imposed a constraint that all the outgoing kernels from a feature map must have the same pruning mask. In this work, we do not impose any such constraint and the pruning granularities are coarser. We argue that this kind of sparsity is useful for VLSI and FFT based implementations. Moreover we show that the best pruning results are obtained when we combine feature map and kernel level pruning. The selection of feature map and kernel pruning candidates with a simple technique is another contribution of this work. 

The rest of the paper is organized as follows. In Section~\ref{sec_relatedWorks}, recent related works are revisited. Section~\ref{sec_candidate} discusses the pruning candidate selection. Section~\ref{sec_fmap_ker} discusses the two pruning granularities. Section~\ref{sec_experiments} presents the experimental results while Section~\ref{sec_conclusion} concludes the discussion and adds the future research dimensions of this work.

\section{Related work}
\label{sec_relatedWorks}

In the literature, network pruning has been studied by several researches \cite{han2015learning, han2015deep, yu2012exploiting, castellano1997iterative,collins2014memory,stepniewski1997pruning,reed1993pruning}. \cite{collins2014memory} have proposed a technique where irregular sparsity is used to reduce the computational complexity in convolutional and fully connected layers. However they have not discussed how the sparse representation will affect the computational benefits. The works of \cite{han2015learning,han2015deep} introduce fine-grained sparsity in a network by pruning scalar weights. If the absolute magnitude of any weight is less than a scalar threshold, the weight is pruned. This work therefore favours learning with small valued weights and train the network with the L1/L2 norm augmented loss function.  Due to pruning at very fine scales, they achieve excellent pruning ratios. However this kind of pruning results in irregular connectivity patterns and demand complex sparse representation for computational benefits. Convolutions are unrolled to matrix-matrix multiplication in \cite{chellapilla2006high} for efficient implementation. The work of \cite{lebedev2015fast} also induce intra-kernel sparsity in a convolutional layer. Their target is efficient computation by unrolling convolutions as matrx-matrix multiplication. Their sparse representation is not also simple because each kernel has an equally sized pruning mask. A recently published work propose sparsity at a higher granularity and induce channel level sparsity in a CNN network for deep face application \cite{polyak2015channel}. The work of  \cite{castellano1997iterative,collins2014memory,stepniewski1997pruning,reed1993pruning} utilize unstructured fine grained sparsity in a neural network. Fixed point optimization for deep neural networks is employed by \cite{anwar2015fixed, hwang2014fixed, sungresiliency} for VLSI based implementations.

\section{Pruning candidate selection}
\label{sec_candidate}

The learning capability of a network is determined by its architecture and the number of effective learnable parameters. Pruning reduces this number and inevitably degrades the classification performance. The pruning candidate selection is therefore of prime importance. Further the pruned network is retrained to compensate for the pruning losses \cite{yu2012exploiting}. For a specific pruning ratio, we search for the best pruning masks which afflicts the least adversary on the pruned network. Indeed retraining can partially or fully recover the pruning losses, but the lesser the losses, the more pluasible is the recovery. Further small performace degradation also means that the successor student network has lost little or no knowledge of the predecessor teacher network. If there are \textit{M} potential pruning candidates, the total number of pruning masks is $(2^{M})$ and an exhaustive search is therefore infeasible even for a small sized network. We therefore propose a simple strategy for selecting pruning candiates. 

\begin{figure}
\centering
\begin{subfigure}{.5\textwidth}
  \centering
  \includegraphics[width=.95\linewidth]{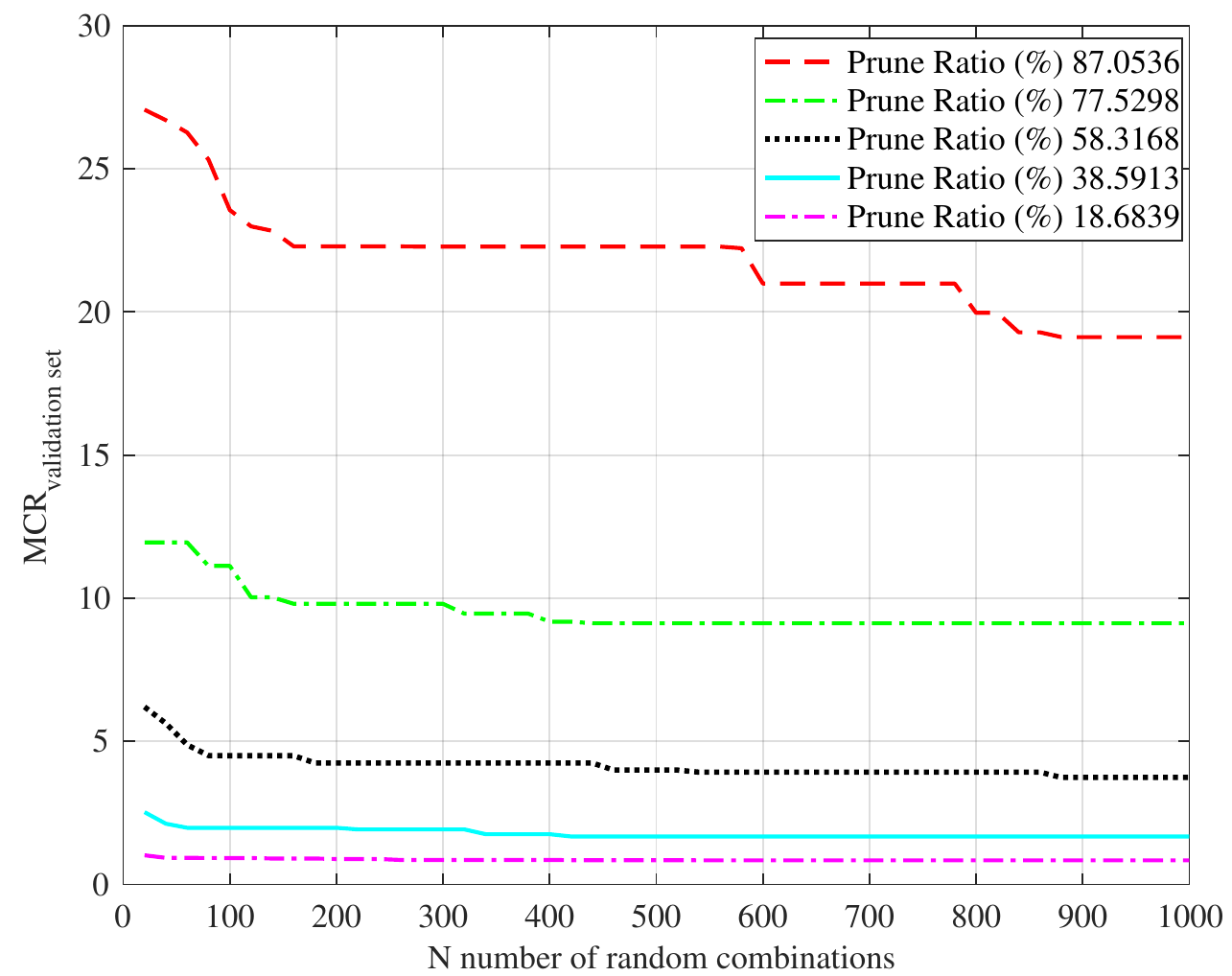}
  \caption{Best of \textit{N} random masks}
  \label{randsearch}
\end{subfigure}%
\begin{subfigure}{.5\textwidth}
  \centering
  \includegraphics[width=.82\linewidth]{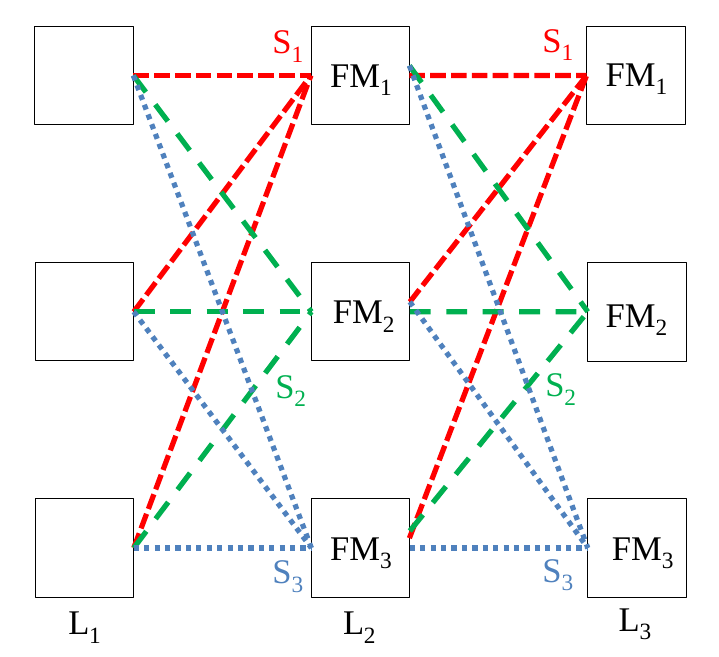}
  \caption{Selecting pruning candidates with weight sum}
  \label{sum_vote}
\end{subfigure}
\caption{(a) MNIST for the network architecture: $1\times6(C5)-16(C5)-120(C5)-84-10$. This figure compares the best candidate selected out of \textit {N} random combinations for various pruning ratios. This figure is plotted for kernel level pruning but is equally applicable to feature map pruning. (b)  This Figure explains the idea presented in \cite{li2016pruning} and shows three layers, $L1$, $L2$ and $L3$. All the filters/kernels from previous layer to a feature map constitute one group which is shown with similr color. The $S1$,$S2$ and $S3$ is computed by summing the absolute value of all the weights in this group.}
\label{fig:rand_sum}
\end{figure}

We evaluate \textit{N} random combinations and compute the MCR for each one. We then choose the best pruning mask which causes the least degradation to the network performance on the validation set. Consider that for the depicted architecture in Fig.\ref{sum_vote}, we need to select feature map pruning candidates in Layer $L_{2}$ and $L_{3}$ with $1/3$ pruning ratio. If \textit{N = 4}, the following \textit{N} ordered pairs of feature maps may be randomly selected for ($L_{2}$, $L_{3}$) : (1, 2), (2, 3), (3, 1), (1, 1). These combinations generate random paths in the network and we evaluate the validation set MCR through these routes in the network. However, this further raises the question of how to approximate \textit{N}. We report the relationship between pruning ratio and \textit{N} in Fig. \ref{randsearch}. This analaysis is conducted for kernel level pruning but is also applicable to feature map level pruning. From Fig. \ref{randsearch}., we can observe that for higher pruning ratios, high value of \textit{N} is beneficial as it results in better pruning candidate selection. For the pruning ratio of no more than 40\%, \textit{N = 100} random evaluations generate good selections. For lower pruning ratios, retraining is also more likely to compensate the losses as the non-pruned parameters may still be in good numbers. The computational cost of this technique is not much as the evaluation is done on the small sized validation set. By observing Fig. \ref{randsearch}., we propose that the value of \textit{N} can be estimated intially and later used in several pruning passes. 
%The pruning ratios were computed for conv2 and conv3 as the layerwise sensitivity analysis shown in \ref{PruningSensitivity_Conv1-2-3}. 

%\begin{figure}[!t]
%\centering
%\includegraphics[width=0.5\columnwidth]{PruningSensitivity_Conv1-2-3}
%\caption{This figure shows the layer wise pruning sensitivity analysis for Conv1,2 and 3. The experiment is conducted with MNIST for the network architecture: $1\times6(C5)-6\times16(C5)-16\times120(C5)-84-10$. Conv1 is most sensitive to pruning due to few number of parameters.}
%\label{PruningSensitivity_Conv1-2-3}
%\end{figure} 
%
%\begin{figure}[!t]
%\centering
%\includegraphics[width=0.5\columnwidth]{randsearch3}
%\caption{This figure compares the best candidate selected out of \textit {N} random combinations for various pruning ratios. The experiment is conducted with MNIST for the network architecture: $1\times6(C5)-6\times16(C5)-16\times120(C5)-84-10$. This figure is plotted for kernel level pruning but is equally applicable to feature map pruning.}
%\label{randsearch}
%\end{figure} 

We further explain and compare this method with the weight sum criterion proposed in \cite{li2016pruning} and shown in Fig. \ref{sum_vote}. The set of filters or kernels from the previous layer constitute a group. This is shown with the similar color in Fig. \ref{sum_vote}. According to \cite{li2016pruning}, the absolute sum of weights determine the importance of a feature map. Suppose that in Fig.\ref{sum_vote}, the Layer $L_{2}$ undergoes feature map pruning. The weight sum criterion computes the absolute weight sum at $S_{1}$, $S_{2}$ and $S_{3}$. If we further suppose that the pruning ratio is $1/3$, then the $min(S_{1}, S_{2},S_{3})$ is pruned. All the incoming and outgoing kernels from the pruned feature map are also removed. We argue that the sign of a weight in kernel plays important role in well-known feature extractors and therefore this is not a good criterion.  

%For coarse pruning granularities, we argue(replace) that a scalar threshold \cite{han2015learning} is not appropriate. In Fig. \ref{magnitude_based}, we show two $3\times 3$ kernerls: \textit{a} shows a Sobel edge detection operator while \textit{b = abs(a)}. If we simply compute the sum of all the weights in both matrices, \textit{a} will be pruned while  \textit{b} will be retained. If we consider the absolute magnitude, similar importance is given to \textit{a} and \textit{b}. However in practice \textit{a} is a more useful edge detector and the sign of a weight is important. We therefore generate \textit{N} random combinations for a specific pruning ratio. We evaluate the network performance with each of the \textit{N} combinations and choose the one with the least adversary on the pruned network performance. 

%\begin{figure}[!t]
%\centering
%\includegraphics[width=0.8\columnwidth]{weight_sum_voting_idea}
%\caption{This Figure explains the idea presented in \cite{li2016pruning} and shows three layers, L1, L2 and L3. All the filters/kernels from previous layer to a feature map constitute one group which is shown with similr color. The S1,S2 and S3 is computed by summing the absolute value of all the weights in this group. }
%\label{sum_vote}
%\end{figure} 

%\begin{figure}[!t]
%\centering
%\includegraphics[width=0.8\columnwidth]{magnitude_based}
%\caption{Figure shows two $3\times3$ kernels. The weight sum absolute method gives equal weightage to \textit{a} and \textit{b} which is not fair.}
%\label{magnitude_based}
%\end{figure} 

\begin{figure}
\centering
\begin{subfigure}{.5\textwidth}
  \centering
  \includegraphics[width=.90\linewidth]{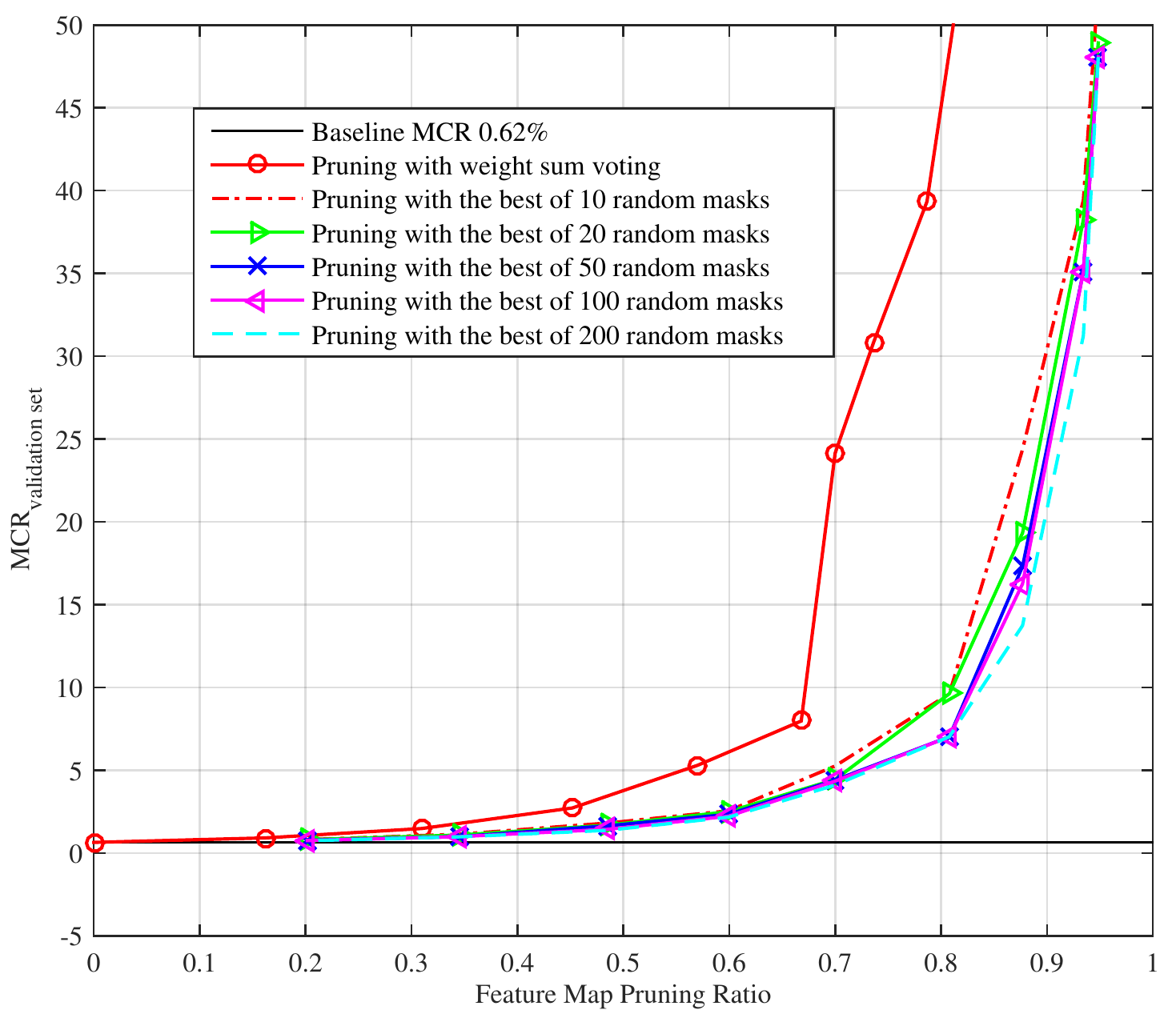}
  \caption{Best of $N$ random masks vs \cite{li2016pruning}}
  \label{weightsum_vs_randsearch}
\end{subfigure}%
\begin{subfigure}{.5\textwidth}
  \centering
  \includegraphics[width=.95\linewidth]{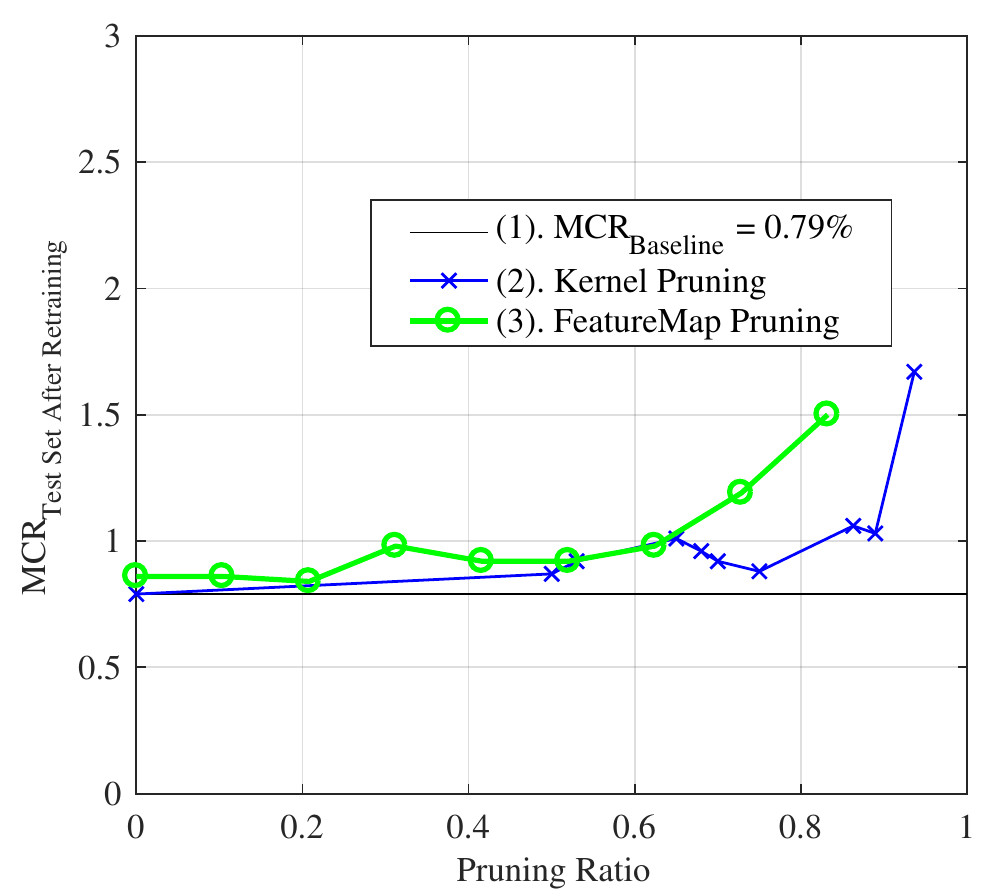}
  \caption{MNIST feature map and kernel pruning}
  \label{mnist_only_kernel_and_fmap}
\end{subfigure}
\caption{(a)This figure compares the feature map pruning candidate selection with weight sum voting and selecting the best out of \textit {N} random combinations for various pruning ratios. The experiment is conducted with the MNIST network architecture: $1\times16(C5)-16\times32(C5)-32\times64(C5)-120-10$. (b)  Figure Feature map and kernel  level pruning applied to the MNIST network. the architecture: $1\times6(C5)-6\times16(C5)-16\times120(C5)-84-10$.}
\label{fig:rand_sum}
\end{figure}

We compare the performance of the two algoirthms and Fig. \ref{weightsum_vs_randsearch} shows the experimental results. These results present the network status before any retraining is conducted. We report the performance degradation in the network classifcation against the pruning ratio. From  Fig.~\ref{weightsum_vs_randsearch}, we can observe that our proposed method outperforms the weight sum method particularly for higher pruning ratios.
This is attributed to evaluating pruning candidates in combinations. The criterion in \cite{li2016pruning} evaluates the importance of a pruning unit in isolation while our proposed approach evaluates several paths through the network and selects the best one. The combinations work togehter and matter more instead of individidual units. Further, our proposed technique is generic and can be used for any pruning granularity: feature map, kernel and intra-kernel pruning. 

\section{Feature map and kernel pruning}
\label{sec_fmap_ker}

% feature map pruning
In this section we discuss feature map and kernel pruning granularities. For a similar sized network, we analyse the achievable pruning ratios with feature map and kernel pruning. In terms of granularity, feature map pruning is coarser than kernel pruning. Feature map pruning does not need any sparse representation and the pruned network can be implemented in a conventional way, convolution lowering \cite{chellapilla2006high} or convolution with FFTs \cite{mathieu2013fast}. The main focus of the proposed work is analysing the unconstrained kernel pruning and feature map pruning. Pruning a feature map causes all the incoming and outgoing kernels to be zeroed because the outgoing kernels are no more meaningful.  

% Kernel level pruning
Kernel pruning is comparatively finer. The dimension and connectivity pattern of 2D kernels determine the computing cost of a convolutional layer. The meshed fully connected convolution layers increases this cost and can hinder the real-time inference. The unconstrained kernel pruning converts this dense connectivity to sparse one. Kernel-pruning zeroes $k \times k$ kernels and is neither too fine nor too coarse. Kernel level pruning provides a balance between fine graind and coarse graind pruining. It is coarse than intra-kernel sparsity and finer than feature map pruning. Thats why good pruing ratios can be achieved at very small sparse representation and computational cost. Each convolution connection represents one convolution operation which involves $Width \times Height \times k \times k$ MAC operations. In LeNet \cite{lecun1998gradient}, the second convolution layer has $6\times16$ feature maps and the kernel connectivity has a fixed sparse pattern. With kernel pruning, we learn this pattern and achieve the best possible pruning ratios. We first select pruning candidates with the criterion outlined in Section 2.  The pruned network is then retrained to compensate for the losses incurred due to pruning. Figure \ref{mnist_only_kernel_and_fmap} shows the feature map and kernel level pruning applied to MNIST \cite{lecun1998gradient} network. When pruning ratios increase beyond 60\%, feature map pruning degrades the performance much. However the kernel level pruning can achieve higher pruning ratios due to finer scale granularity.

%\begin{figure}[!t]
%\centering
%\includegraphics[width=0.8\columnwidth]{kernel_prune_idea}
%\caption{The $2 \times 2 = 4$ convolution connections are reduced to $2$ convolutions. Thus the dense connectivity pattern is now sparse one. Pruning the kernels with high ratios may eliminate feature maps if all the incoming kernels are zeroed.}
%\label{kernel_prune_idea}
%\end{figure} 

As the sparse granularities are coarse, a generic set of computing platform can benefit from it. One downside of the unconstrained kernel pruning is that convolutions can not be unrolled as matrix-matrix multiplications \cite{chellapilla2006high}. However, customized VLSI implementations and FFT based convolutions do not employ convolution unrolling. Mathieu et. al., have proposed FFT based convolutions for faster CNN training and evaluation~\cite{mathieu2013fast}. The GPU based parallel implementation showed very good speedups. As commonly known that the $IFFT(FFT(kerenel)~\times~FFT(fmap)) = kernel \ast fmap$, the kernel level pruning can relieve this task.  Although the kernel size is small, massive reusability enables the use of FFT. The FFT of each kernel is computed only once and reused for multiple input vectors in a mini-batch. In a feed forward and backward path, the summations can be carried in the FFT domain and once the sum is available, the IFFT can be performed \cite{mathieu2013fast}.  Similarly, a customized VLSI based implementation can also benefit from the kernel level pruning. If the VLSI implementation imposes a constraint on the pruning criterion, such as the fixed number of convolution kernels from the previous to the next layer, the pruning criterion can be adapted accordingly. Figure \ref{mnist_only_kernel_and_fmap} shows that the kernel pruning can be induced in much higher rates with minor increase in the MCR of the baseline MNIST network. In the next Section, we report and discuss the experimental results in detail. 

\section{Experimental Results}
\label{sec_experiments}

% \usepackage{graphicx}

In this section, we present detailed experimental results with the CIFAR-10 and SVHN datasets \cite{krizhevsky2009learning}.  During training and pruning, we use the stochastic gradient descent (SGD) with a mini-batch size of 128 and RMSProp \cite{tieleman2012lecture}. We train all the networks with batch normalization \cite{ioffe2015batch}. We do not prune the network in small steps, and instead one-shot prune the network for a given pruning ratio followed by retraining. The experimental results are reported in the corresponding two subsections. 

\begin{table*}[t]
\centering
\caption{Specifications of the three CIFAR-10 networks}
\label{Table1}
\renewcommand{\arraystretch}{1.2}
\resizebox{\textwidth}{!}{%
\begin{tabular}{ccclll}
%\multicolumn{3}{c}{Table 1. Specifications of the three CIFAR-10 networks}                                                                                         %&  &  &  
\\ \cline{1-3}
\multicolumn{1}{|c|}{Network}           & \multicolumn{1}{c|}{Architecture}                                                    & \multicolumn{1}{c|}{Baseline MCR(\%)} &  &  &  \\ \cline{1-3}
\multicolumn{1}{|c|}{$CNN_{small}$}     & \multicolumn{1}{c|}{2x128C3-MP2-2x128C3-MP2-2x256C3-256FC-10Softmax}                 & \multicolumn{1}{c|}{16.6}         &  &  &  \\
\multicolumn{1}{|c|}{$CNN_{large}$} & \multicolumn{1}{c|}{2x128C3-MP2-2x256C3-MP2-2x256C3-1x512C3-1024FC-1024FC-10Softmax} & \multicolumn{1}{c|}{9.41}         &  &  &  \\ \cline{1-3}
\multicolumn{1}{l}{}                    & \multicolumn{1}{l}{}                                                                 & \multicolumn{1}{l}{}              &  &  &  \\
\multicolumn{1}{l}{}                    & \multicolumn{1}{l}{}                                                                 & \multicolumn{1}{l}{}              &  &  & 
\end{tabular}%
}
\end{table*}

\subsection{CIFAR-10}
\label{subsec_cifar10}

The CIFAR-10 dataset includes samples from ten classes: airplane, automobile, bird, cat, deer, dog, frog, horse, ship and truck. The training set consists of 50,000 RGB samples and we allocate 20\% of these samples as validation set. Test set contains 10,000 samples and each sample has $32 \times 32 \times RGB$ resolution. We evaluate the proposed pruning granularities with two networks. $CNN_{small}$ and $CNN_{large}$. $CNN_{small}$ has six convolution and two overlapped max pooling layers. We report the network architecture with an alphanumeric string as reported in \cite{courbariaux2015binaryconnect} and outlined in Table~\ref{Table1}. The $\left(2\times128C3\right)$ represents two convolution layers with each having 128 feature maps and $3\times 3$ convolution kernels. $MP2$ represents $3\times 3$ overlapped max-pooling layer with a stride size of 2.  We pre-process the original CIFAR-10 dataset with global contrast normalization followed by zero component analysis (ZCA) whitening. 

The $CNN_{large}$ has seven convolution and two max-pooling layers. Furthe, online data augmentations are employed to improve the classification accurracy. We randomly crop $28\times28\times3$ patches from the $32\times32\times3$ input vectors. These cropped vectors are then geometrically transformed randomly. A vector may be flipped horizontally or vertically, rotated, translated and scaled. At evaluation time, we crop patches from the four corners and the center of a $32\times32\times3$ patch and flip it horizontally. We average the evaluation on these ten $28\times28\times3$ patches to decide the final label. Due to larger width and depth, the $CNN_{large}$ achieves more than 90\% accurracy on the CIFAR-10 dataset. The $CNN_{small}$ is smaller than $CNN_{large}$ and trained without any data augmentation. The $CNN_{small}$ therefore achieves 84\% accurracy.

\subsubsection{Feature map and kernel level pruning}
\label{subsubsec_fmap_kernel}

\begin{figure}
\centering
\begin{subfigure}{.5\textwidth}
  \centering
  \includegraphics[width=.9\linewidth]{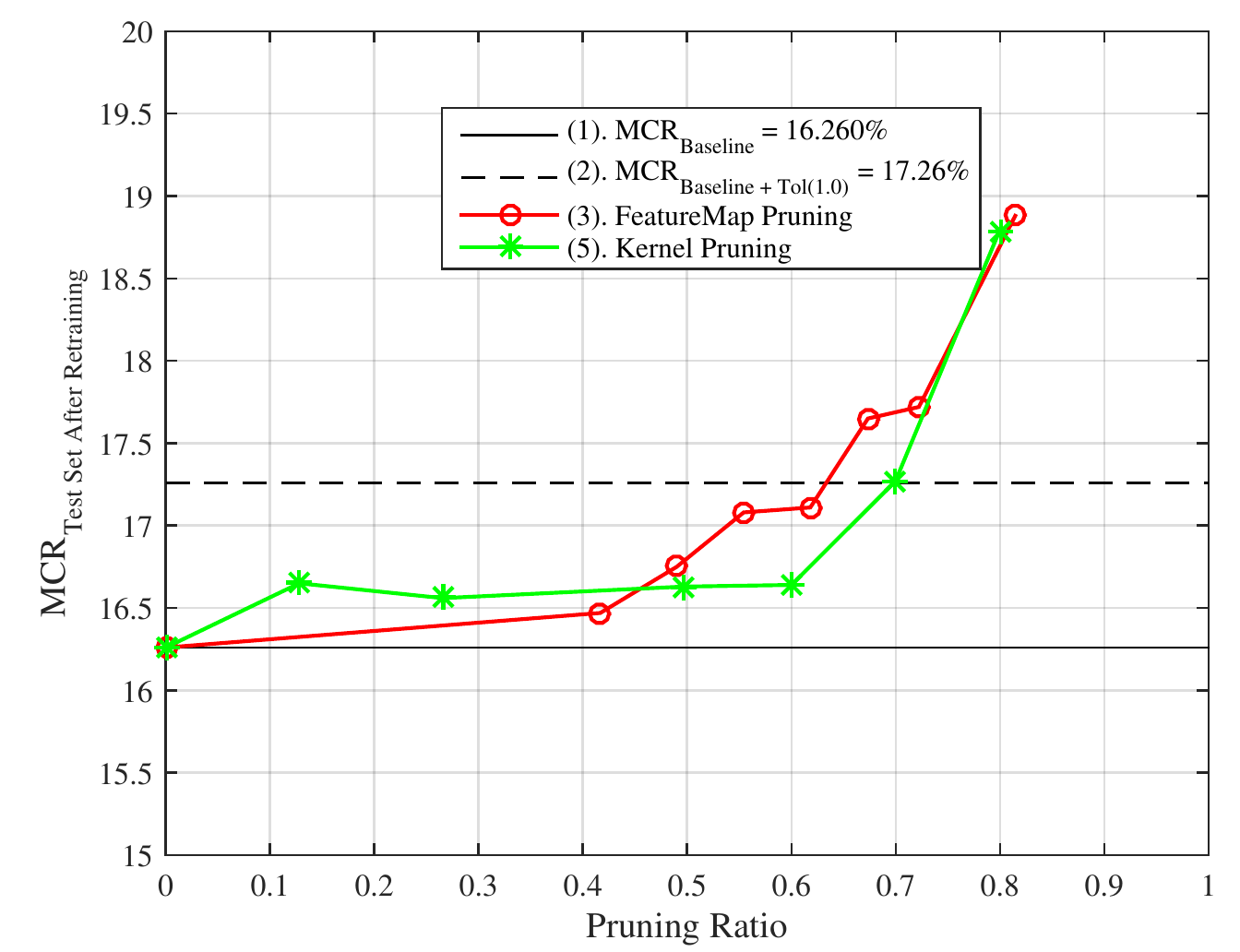}
  \caption{$CNN_{small}$ kernel and feature map pruning}
  \label{cifar10_1}
\end{subfigure}%
\begin{subfigure}{.5\textwidth}
  \centering
  \includegraphics[width=.85\linewidth]{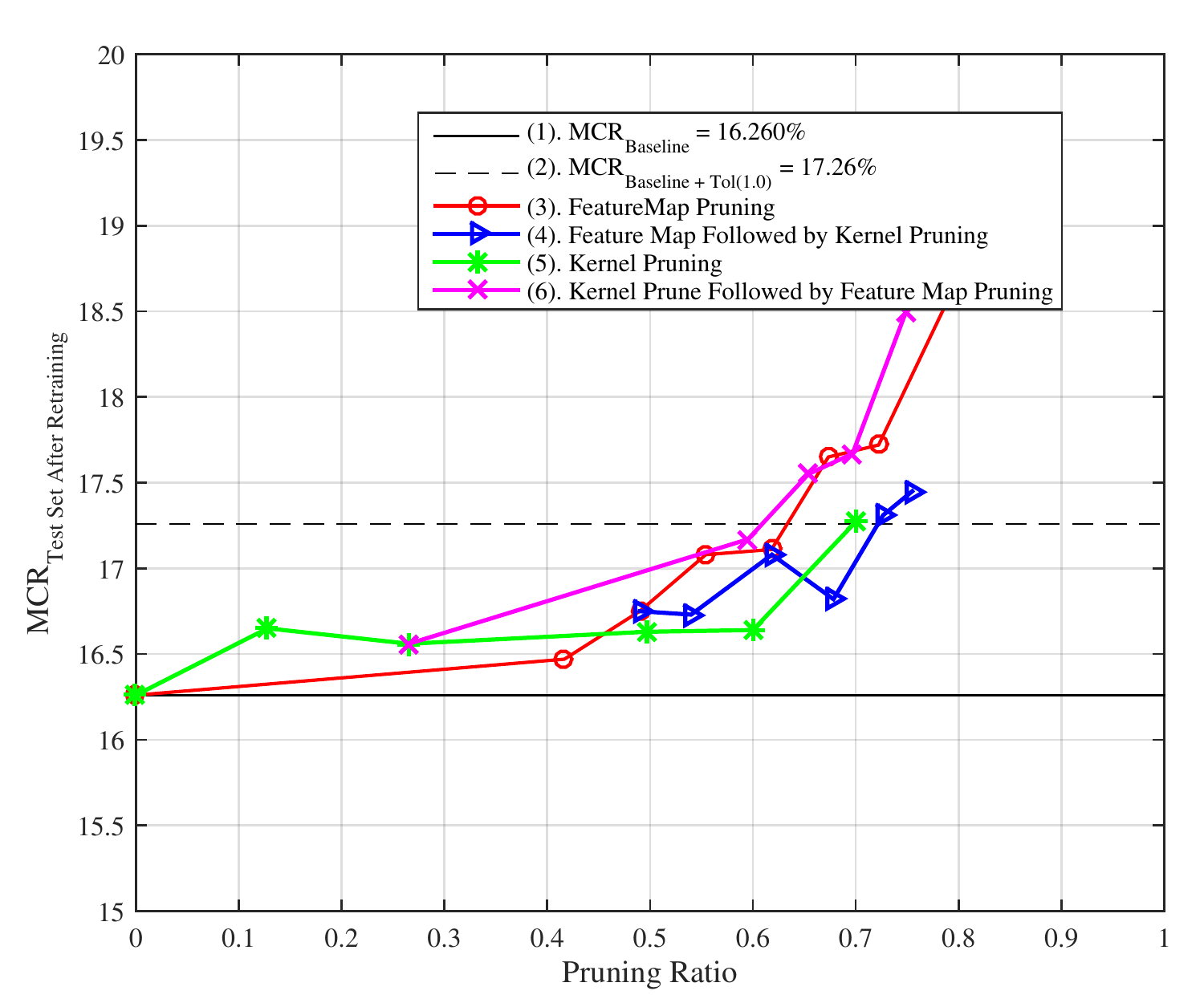}
  \caption{$CNN_{small}$ kernel and feature map pruning applied in various combinations.}
\label{kernel_fmap_and_combinations}
  \label{kernel_fmap_and_combinations}
\end{subfigure}
\caption{(a)This figure prunes the $CNN_{small}$ network with feature maps and kernels. (b) Here we show that higher pruning ratios can be achieved if we apply pruning granularities in various combinations.}
\label{kernel_fmap_and_combinations}
\end{figure}

%\begin{figure}[!t]
%\centering
%\includegraphics[width=0.8\columnwidth]{kernel_and_fmap_only_pruning_2}
%\caption{CIFAR-10 kernel and feature map pruning}
%\label{cifar10_1}
%\end{figure} 
%
%\begin{figure}[!t]
%\centering
%\includegraphics[width=0.8\columnwidth]{kernel_fmap_and_combinations}
%\caption{ This Figure shows pruning applied in various combinations to the CIFAR-10 $CNN_{small}$.}
%\label{kernel_fmap_and_combinations}
%\end{figure} 

After layer pruning, feature map pruning is the 2nd coarsest pruning granularity. Feature map pruning reduces the width of a convolutional layer and generates a thinner network. Pruning a single feature map, zeroes all the incoming and outgoing weights and therefore, higher pruning ratios degrade the network classification performance significantly. Feature map pruning for the $CNN_{cifar}$ is shown in Fig. \ref{cifar10_1} with a circle marked red colored line. The sparsity reported here is for Conv2 to Conv6. We do not pruned the first convolution layer as it has only $3 \times 128 \times (3 \times 3) = 3456$ weights. The horizontal solid line shows the baseline MCR of 16.26\% whereas the dashed line shows the 1\% tolerance bound. Training the network with batch normalization \cite{ioffe2015batch} enables us to directly prune a network for a target ratio, instead of taking small sized steps. With a baseline performance of 16.26\%, the network performance is very bad at 80\% feature map pruning. We can observe that 62\% pruning ratio is possible with less than 1\% increase in MCR. The $CNN_{cifar}$ is reduced to $\left(128C3-83C3\right)$-MP3-$\left(83C3-83C3\right)$-MP3-$\left(166C3-166C3\right)$-$256FC$-$10 Softmax$. As pruning is only applied in Conv2 to Conv6, therefore the Figure \ref{cifar10_1} pruning ratios are computed only for these layers. 

For the same network, we can see that kernel level pruning performs better. We can achieve 70\% sparsity with kernel level pruning. This is attributed to the fact that kernel pruning is finer and hence it achieves higher ratios. Further kernel pruning may ultimately prune a feature map if all the incoming kernels are pruned. However at inference time, we need to define the kernel connectivity pattern which can simply be done with a binary flag. So although the sparse representation is needed, it is quite simple and straightforward. Experimental results confirm that fine grained sparsity can be induced in higher rates. We achieved 70\% kernel wise sparsity for Conv2 - Conv6 and the best pruned network is layer wise reported in Table \ref{kernelprunetable}. The speedup and acceleration with these pruning granularities is platform independent.
%\begin{table}[]
%\centering
%\caption{Kernel level pruning (70\%) in $CNN_{small}$}
%\label{kernelprunetable}
%\begin{tabular}{|cccc|}
%\hline
%Conv Layer        & Kernel Conn & Pruned Conn        & Pruned Ratio \\ \hline
%C2 (128x128) & 16384       & 104139 / 9 = 11571 & 71\%         \\
%C3 (128x128) & 16384       & 102888 / 9 = 11432 & 70\%         \\
%C4 (128x128) & 16384       & 103437 / 9 = 11493 & 70\%         \\
%C5 (128x256) & 32768       & 206172 / 9 = 32768 & 70\%         \\
%C6 (266x256) & 65536       & 412029 / 9 = 45781 & 70\%         \\ \hline
%\end{tabular}
%\end{table}

\begin{table*}[]
\centering
\caption{Feature map and kernel level pruning (75\%) in $CNN_{small}$}
\label{my-label}
\begin{tabular}{|cccccc|}
\hline
Layer & Fmap Pruned & Pruned Kernels & Kernel Conn & \multicolumn{1}{l}{Kernel Prune} &\\ 
Fmaps & To (\%) & $(3 \times 3 = 9)$  & & \multicolumn{1}{l}{Ratio(\%)} &\\ \hline
C2 (128x128) & C2 (128x89), 30.5 & 27306/9 = 3034 & 11392 & 3034/11392 = 26.6 &\\
C3 (128x128) & C3 (89x89), 51.5 & 18702/9 = 2078 & 7921 & 2078/7921 = 26.2 &\\
C4 (128x128) & C4 (89x89), 51.5 & 18702/9 = 2078 & 7921 & 2078/7921 = 26.2  &\\
C5 (128x256) & C5 (89x179), 51.4 & 37881/9 = 4209 & 15931 & 4209/15931 = 26.4  &\\
C6 (266x256) & C6 (179x179), 51.1 & 76851/9 = 8539 & 32041 & 8539/32041 = 26.6 & \\ \hline
\end{tabular}
\end{table*}

\subsubsection{Combinations of Kernel and feature map pruning}

In this section we discuss the various pruning granularities applied in different combinations. We first apply the feature map and kernel pruning to the $CNN_{small}$ network in different orders. With feature map pruning, we can achieve 60\% sparsity under the budget of 1\% increase in MCR. But at this pruning stage, the network learning capability is affected much. So we take a 50\% feature map pruned network, where the $CNN_{small}$ is reduced to $\left(128C3-89C3\right)$-MP3-$\left(89C3-89C3\right)$-MP3-$\left(179C3-179C3\right)$-$256FC$-$10 Softmax$. As pruning is only applied to $Conv2 - Conv6$, therefore in Fig. \ref{kernel_fmap_and_combinations}., pruning ratios are computed only for these layers. This network then undergoes kernel level pruning. The blue rectange line in Figure \ref{kernel_fmap_and_combinations} shows the pruning results. We achieve the best pruning results in this case and the final pruned network is reported in detail in Table 3. Overall we achieve more than 75\% pruning ratio in the final pruned network.  

\begin{figure}[!t]
\centering
\includegraphics[width=0.8\columnwidth]{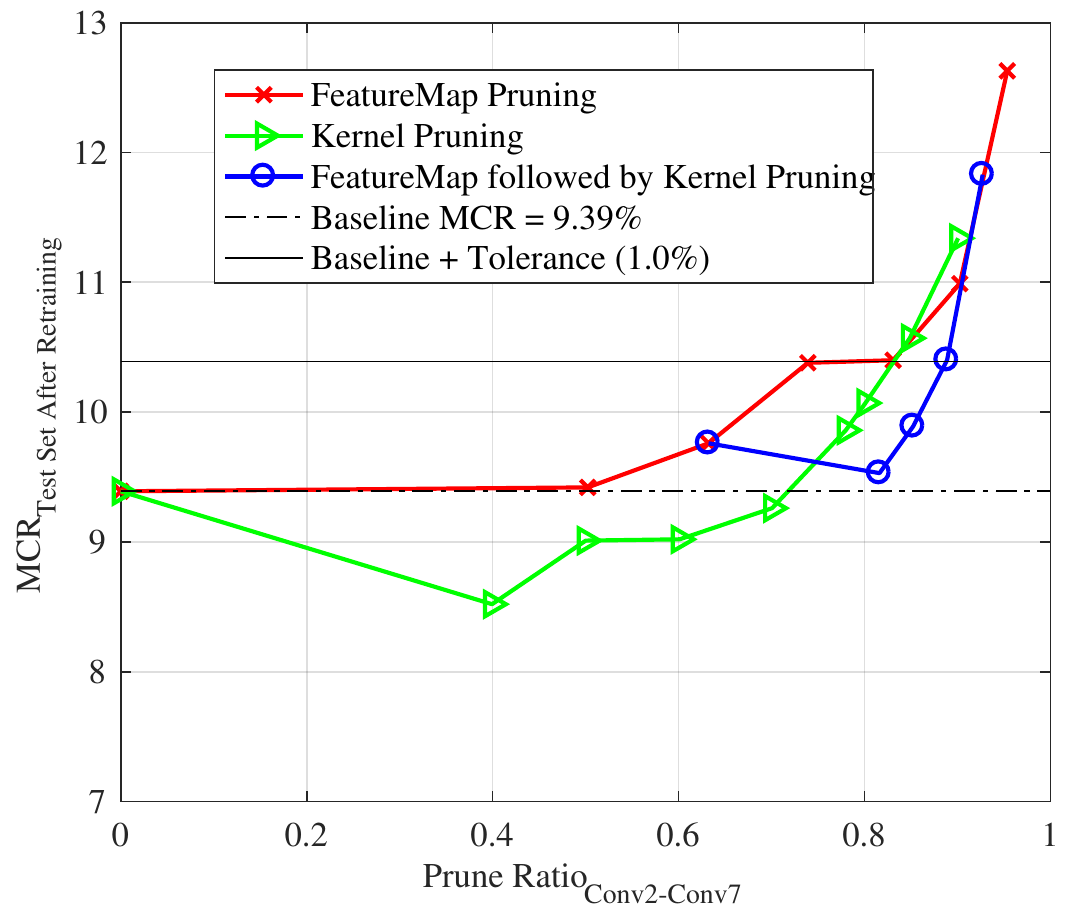}
\caption{The figure shows various pruning combinations applied to the CIFAR-10 $CNN_{large}$.}
\label{cifar_large_2}
\end{figure}

We further conducted experiments on the $CNN_{large}$ and the corresponding plots are shown in Fig. \ref{cifar_large_2}. The $CNN_{large}$ is much wider and deeper than the CNNsmall as reported in Table 1. Therefore there are more chances of redundancy and hence more room for pruning. Further we observe similar trends as $CNN_{small}$ where the kernel pruning can be induced in higher ratios compared to the feature map pruning. When the kernel pruning is applied to the feature map pruned network, we can achieve more than 88\% sparsity in the $Conv2-Conv7$ of the $CNN_{large}$ network. This way we show that our proposed technique has good scalability. These results are in conformity to the resiliency analysis of fixed point deep neural networks \cite{sungresiliency}.

\subsection {SVHN}

\begin{figure}[!t]
\centering
\includegraphics[width=0.8\columnwidth]{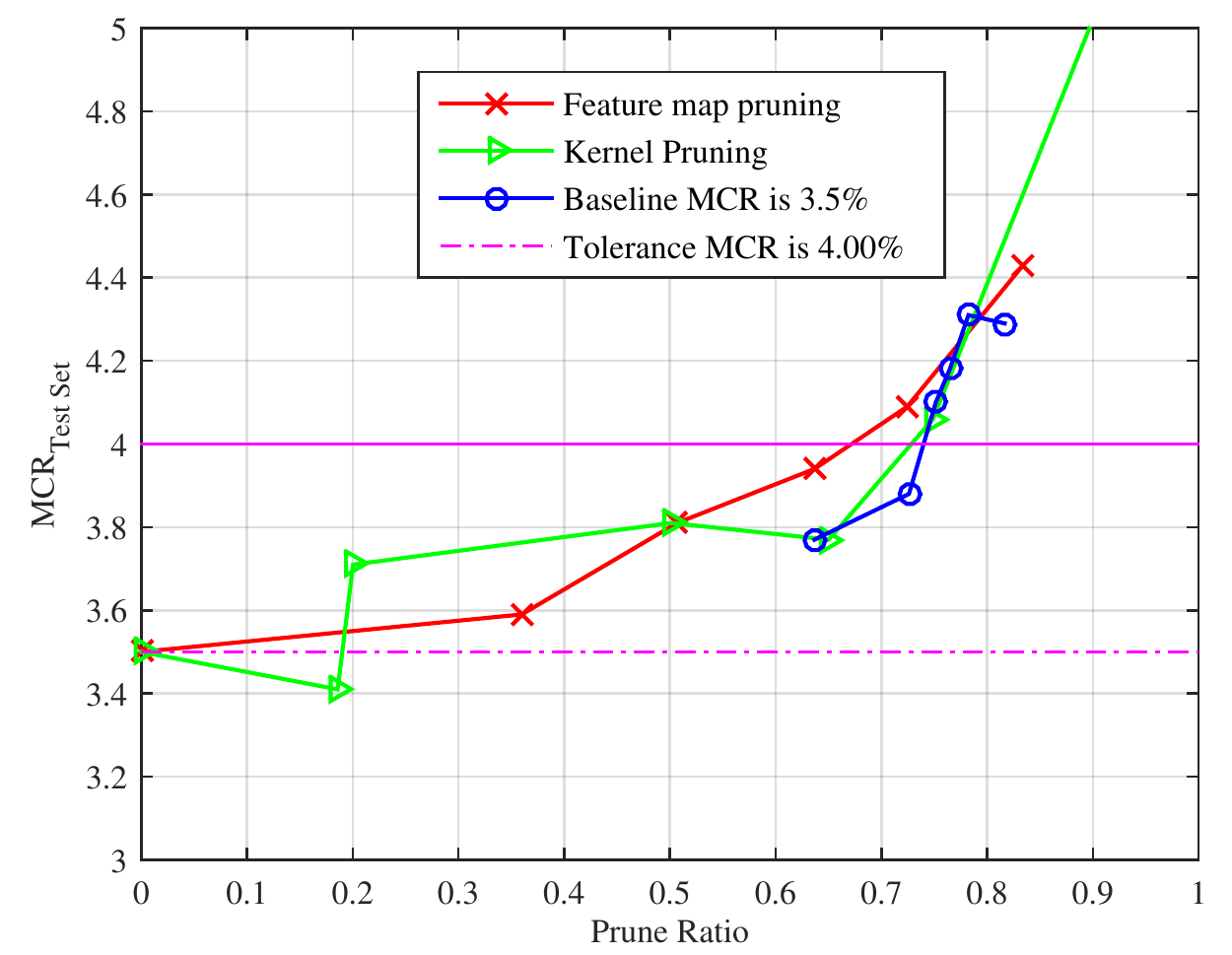}
\caption{ This Figure shows pruning applied in various combinations to the SVHN CNN network.}
\label{svhn_2}
\end{figure}

The SVHN dataset consists of $32\times32\times3$ cropped images of house numbers [Netzer et al.
2011] and bears similarity with the MNIST handwritten digit recognition dataset [LeCun et al. 1998]. 
The classification is challenging as more than one digit may appear in sample and the goal is to identify a digit in the center of a patch.
The dataset consists of 73,257 digits for training, 26,032 for testing and 53,1131 extra for training. 
The extra set consists of easy samples and may augment the training set. We generate a validation set of 6000 samples which consists of 4000 samples from the training set and 2000 samples from the extra [Sermanet
et al. 2012]. The network architecture is reported like this: $(2\times64C3)$-MP2-
$(2\times128C3)$-MP2-$(2\times128C3)$-512FC-512FC-10Softmax. This network is trained with
batch normalization and we achieve the baseline MCR of 3.5\% on the test set. The corresponding pruning plots are reported in Fig. \ref{svhn_2}. 
We can observe a similar trend where kernels can be pruned by a bigger ratio compred to feature maps. More than 70\% pruning ratio can be implemented in the reported network.
Thus we show that the lessons learnt generalize well on various datasets. 

\section{Concluding Remarks}
\label{sec_conclusion}

In this work we proposed feature map and kernel pruning for reducing the computational complexity of deep CNN. We have discussed that the cost of sparse representation can be avoided with coarse pruning granularities. We demonstrated a simple and generic algorithm for selecting the best pruning mask. We conducted experiments with several benchmarks and networks and showed that the proposed technique has good scalability. We are exploring online pruning in future for exploiting run-time benefits.

\section*{Acknowledgment}

This work was supported by the National Research Foundation of Korea (NRF) grant funded by the Korean government (MSIP) (No. 2015R1A2A1A10056051).

\bibliography{iclr2017_conference}
\bibliographystyle{iclr2017_conference}

\end{document}